\def\year{2021}\relax
\documentclass[letterpaper]{article} 
\usepackage{aaai21}  
\usepackage{times}  
\usepackage{helvet} 
\usepackage{courier}  
\usepackage[hyphens]{url}  
\usepackage{graphicx} 
\usepackage{natbib}  
\usepackage{caption} 
\usepackage{xcolor}
\usepackage{hyperref}

\definecolor{myred}{rgb}{0.84,0.1,0.1}

\title{Lifelong Learning Dialogue Systems: Chatbots that Self-Learn On the Job}

\author {
     Bing Liu,~ 
     Sahisnu Mazumder \\ 
 }
 \affiliations {
     Department of Computer Science, University of Illinois at Chicago, USA \\
     liub@uic.edu,~ sahisnumazumder@gmail.com 
 }

\begin{document}
\maketitle

{\large \color{myred}  An \textbf{extended and revised version} of this work has been \textbf{published} in the proceedings of AAAI as follows: 

\vspace{2mm}	
\begin{quote}
	 Bing Liu and Sahisnu Mazumder. \textbf{Lifelong and Continual Learning Dialogue Systems:  Learning during Conversation}. \textit{In Proceedings of the AAAI Conference on Artificial Intelligence (AAAI), 2021.}. 
	 
	 \vspace{1mm}	
	 click here: {\color{blue} \href{https://www.cs.uic.edu/~liub/publications/LINC_paper_AAAI_2021_camera_ready.pdf}{Camera-ready version link}}
	 
	Official version will appear on Web soon.
\end{quote}		

\noindent
\underline{\textbf{Please consider the AAAI-2021 version for citation}}\\ \underline{\textbf{as mentioned above.}} \\
}

\begin{abstract}
Dialogue systems, also called \textit{chatbots}, are now used in a wide range of applications. However, they still have some major weaknesses. One key weakness is that they are typically trained from manually-labeled data and/or written with handcrafted rules, and their knowledge bases (KBs) are also compiled by human experts. Due to the huge amount of manual effort involved, they are difficult to scale and also tend to produce many errors ought to their limited ability to understand natural language and the limited knowledge in their KBs. Thus, the level of user satisfactory is often low. In this paper, we propose to dramatically improve this situation by endowing the system the ability to continually learn (1) new world knowledge, (2) new language expressions to ground them to actions, and (3) new conversational skills, during conversation or \textit{on the job} by themselves so that as the systems chat more and more with users, they become more and more knowledgeable and are better and better able to understand diverse natural language expressions and improve their conversational skills. A key approach to achieving these is to exploit the multi-user environment of such systems to self-learn through interactions with users via verb and non-verb means. The paper discusses not only key challenges and promising directions to learn from users during conversation but also how to ensure the correctness of the learned knowledge.






\end{abstract}

\section{Introduction}
Building \textit{dialogue systems} or \textit{conversational agents} capable of conversing with humans in natural language (NL) and understanding human NL instructions is a long-standing goal of AI \cite{winograd1972understanding}. These agents, also called \textit{chatbots}, have become the front runner of AI advancement due to wide-spread applications 
such as assisting customers in buying products, booking flight tickets, reducing stress, and executing actions 
like controlling house appliances and reporting weather information. 
Because of the proliferation of Internet of Things (IoT) with NL interfaces, the importance of these agents have become ubiquitous in recent times. 

Conversational agents can be broadly categorized into two main types: 
\textbf{(1) Chit-chat systems} \cite{shang2015neural,sordoni2015neural,li2016deep,serban2016building,serban2017hierarchical} designed to engage users and provide mental support by conducting chit-chat type of conversation in wide range of topics without having a specific goal to complete. \textbf{(2) Task-oriented chatbots} \cite{raux2005let,williams2007partially,wen2017network} designed to assist users to complete tasks based on users' requests, e.g., providing the requested information and taking actions. Most of the popular personal assistants such as Amazon Alexa, Apple Siri, Google Home, and Microsoft Cortana, are task-oriented bots. They are primarily designed as Natural Language Interaction (NLI) systems that take human NL instructions (commands) and translate them into some actions to be executed by the underlying application. Question-answering (QA) and conversational recommendation systems also fall into this category.
\footnote{This paper is not concerned with speech-to-text conversion.}

Before deep learning became popular, chatbots were developed mainly using the markup language AIML\footnote{http://www.alicebot.org/} or handcrafted conversation generation rules. With the advent of deep learning, the trend has shifted toward end-to-end conversation modeling \cite{vinyals2015neural,xing2017topic,wen2017network}. However, despite the fact that chatbots are widely used, they still have some serious weaknesses: \textbf{(1)} A great deal of manual effort is needed to label training data, write rules and compile knowledge bases (KBs). No matter how much data is collected and used to train a chatbot, it is hard to cover all possible variations of natural language. Thus, when deployed in practice, a well-trained chatbot often performs poorly. \textbf{(2)} The pre-compiled KBs cannot cover the rich knowledge needed in practice.  

This paper argues that a truly intelligent chatbot should not be limited by its offline-trained model or pre-compiled KB. 
It should learn continuously \textit{on the job}, i.e., after model deployment and during conversing or interacting with the (human) end users and thereby, improve its capability over time in a self-supervised manner~\cite{ChenAndLiubook2018,liu2020learning}. 
Thus, this paper proposes the new paradigm, called \textbf{\underline{L}ifelong \underline{IN}teractive learning in \underline{C}onversation} (LINC). LINC needs a new definition because traditional lifelong learning is for offline learning of a sequence of tasks with given tasks and given labeled data~\cite{ChenAndLiubook2018}. However, learning during conversation is like human on-the-job learning where the system has to discover its own tasks and also the training data. 
This paper focuses on three continuous learning capabilities of chatbots: (1) learning factual knowledge in open-ended and information-seeking conversations, 
(2) learning to ground new NL commands (language expressions),
and (3) learning new conversational skills from users. Some preliminary work has been done on (1) and (2) in~\cite{mazumder2018towards,mazumder2019building,mazumder2019lifelong}.

The key idea for solving the LINC problem is to exploit the wisdom of the crowd in the multi-user environment in which almost all chatbots work to learn new knowledge by asking or interacting with the current user and/or the other users to enable the chatbot to learn quickly and effectively. 
This powerful approach, however, also comes with a major shortcoming. The knowledge learned from end-users can be erroneous and some users may even purposely fool the system by providing wrong information or knowledge. We will discuss how to solve this problem to ensure the credibility or trustworthiness of the learned knowledge from end users. 
Note, we use the terms: chatbot, bot, agent, NLI systems, dialogue systems interchangeably in the rest of the paper.

\section{The Paradigm of Lifelong Interactive Learning in Conversation (LINC)}
\vspace{0.05cm}
The proposed LINC paradigm is based on the traditional lifelong learning (LL)~\cite{ChenAndLiubook2018,thrun1998lifelong, silver2013lifelong, Ruvolo2013ELLA,chen2014topic}, but also needs significant extensions.
 
 \textbf{Lifelong learning} is stated as follows: At any time point, the learner has performed a sequence of $N$ learning tasks, $T_1$, $T_2$, …, $T_N$ with their corresponding training data $D_1$, $D_2$, …, $D_N$. When faced with the $(N+1)^{th}$ task $T_{N+1}$ with its training data $D_{N+1}$, the learner can transfer the knowledge learned from the previous $N$ tasks to help learn $T_{N+1}$. 


For \textbf{LINC}, this LL definition is insufficient because this definition is for offline learning with given tasks and given training data. But during a conversation, the tasks have to be created by the agent itself on-the-fly and the training data has to be found by it too. A new learning task $T_{N+1}$ is formed when the agent wants to learn a piece of knowledge from a user utterance (e.g., extracting an unknown relation) or encounters a problem in an actual online conversation (e.g., unable to understand a user utterance or unable to answer a user query).\footnote{The knowledge learning tasks created by the agent are entirely different from the tasks the end-user wants to perform via the agent.} In order to learn the new task $T_{N+1}$, it needs to obtain the ground truth training data $D_{N+1}$. That is why the agent has to interact with or ask the user questions in order to obtain the ground truth data and learn from it. This learning process is like human on-the-job learning.



For LINC to succeed, the key challenge is how to obtain the ground truth training data on its own initiative in order to learn the new task $T_{N+1}$. The agent has to: (1) \textit{formulate a dynamic interaction strategy} $\mathcal{I}$ to interact with the user, e.g., deciding what to ask the user and when to ask the user; (2) \textit{execute} $\mathcal{I}$ to acquire the ground truth data; (3) \textit{incrementally learn task} $T_{N+1}$, which is like tradition lifelong/continual learning and will not be discussed further in this paper. 

\vspace{+2mm}
\noindent
\textbf{Interactively Obtain Ground Truth Data.
} As chatbots typically work in a multi-user environment, we propose to exploit such an environment to obtain the ground truth data during an actual online conversation. 


\textbf{1. Extracting information from user utterances:} 
The chatbot can extract information from user utterance (or dialogue history) which can be real-world facts, 
user's preferences etc. Note that, here learning may simply be storing the new information in the KB and inferring additional knowledge from the acquired and existing KB knowledge. 


\textbf{2. Asking the current user:} When the system (1) does not understand a user utterance, or (2) cannot answer a user query, it forms a new learning task. To obtain the ground truth data, for (1), the agent can ask the current user for clarification, rephrasing, or even demonstration if it is supported in the application. For (2), the agent may ask the user for some supporting facts and then infer the query answer. In order to obtain more knowledge, the agent may even ask the current user related questions. For example, the user said ``\textit{I visited London last month.}" Apart from extracting London as a location, it can also ask a subsequent question: ``\textit{Where is London?}" If the user answers ``\textit{London is in UK}," the system learned another piece of knowledge. 

\textbf{3. Asking other users}: When the chatbot could not answer a user query, it may also ask other users to obtain the answer. For example, if a user asks ``\textit{What is the capital city of the US?}" and the agent is not able to answer or infer now, it can try to find a good opportunity in future to ask another user ``\textit{Hey, do you know what the capital city of the US is?}" If the user gives the answer ``\textit{it's Washington DC}," the agent acquires the ground truth (a piece of new knowledge). 

In the next three sections, we discuss the specific problems of learning \textit{factual knowledge}, \textit{natural language expressions}, and \textit{conversation skills} during conversation. 

\section{Factual Knowledge Learning in Conversation}


Many chatbots (e.g., conversational search and question-answering systems) have an explicit knowledge base storing real-world facts [e.g., (\textit{Chicago}, \textit{CityOf}, \textit{USA})] \cite{eric2017key,madotto2018mem2seq,ghazvininejad2017knowledge,le2016lstm,Young2018Augmenting,long2017knowledge,zhou2018commonsense} 
to support information-seeking conversations and help users with product recommendations.
One major issue with existing approaches is that the KBs are fixed once the systems are deployed. However, it is almost impossible for the initial KBs to contain all possible knowledge that the user may ask, not to mention that new knowledge appears constantly. It is thus highly desirable for dialogue systems to acquire new knowledge directly from the user utterances or by explicitly asking users questions while in use. 
\citet{hancock2019learning} proposed a self-feeding chatbot, with the ability to extract new training examples from the conversations. However, they do not focus on interactive factual knowledge learning.

There are many opportunities to learn new knowledge during an actual conversation. Here are a few examples. 

\begin{itemize}
    \item \textbf{Extracting facts from user utterances.}
    For example, while conversing about movies, if the user says ``\textit{I watched Forest Gump yesterday. The movie was awesome. Liked Tom Hanks' performance very much.}", the chatbot can extract new fact (\textit{Forest Gump, isa, movie}) and (\textit{Tom Hanks, performed\_in, Forest Gump}). Later, the chatbot can use these facts in future conversations while answering questions like "\textit{Who acted in Forest Gump?}" or generating a response to user's utterance "\textit{I'm feeling bored. Can you recommend a good movie?}".
    
    \item \textbf{Ask questions to learn about unknown entities and concepts.} As unknown entities and concepts appear frequently in user utterances, the chatbot can ask clarification or information seeking questions to the user to acquire facts about new entities or concepts. For example, if the user says ``\textit{Is there any good place around for having sushi}?", the chatbot can ask, ``\textit{Is sushi a food?}" or ``\textit{what is sushi?}". 
    \citet{otsuka2013generating} and \citet{ono2016toward,ono2017lexical} have explored the problem of lexical acquisition during dialogues in closed-domain chatbots. However, they did not discuss the method in a lifelong learning setting.
    
    \item \textbf{Ask and infer new facts.} As mentioned earlier, when the chatbot cannot answer an user query, it can ask for some related supporting facts and then infer the answer. 
    
    ~~~\citet{mazumder2019lifelong} have studied the problem when the system is unable to answer a user's WH-question, where the system formulates some questions to acquire supporting facts from user and then, uses these facts and existing knowledge in the KB to answer the WH-question. The supporting facts and inferred answers can be regarded as new knowledge (we will discuss how to ensure the correctness of knowledge shortly). The work showed that the performance (Hits@1) improves by 14.4\% if the chatbot allowed to acquire 3 facts per entity from the user compared to that if it only acquires 1 fact. Also, if continuous learning of the prediction model using acquired facts (in past sessions) are disabled, Hits@1 degrades by 7\%. 
\end{itemize}

\noindent Although some existing work explored some of the above opportunities, several challenges are still not addressed:

\textbf{1. Understanding context and topic of conversation:} An entity or concept appearing in a conversation context, can be ambiguous. E.g., ``\textit{apple}" can be a fruit or name of a company. Thus, understanding the topic of conversation or context while grounding facts for response generation is crucial to maintain the relevance of the conversation. 

\textbf{2. Coreference resolution}: In multi-turn dialogues, user can often use coreferences to denote an entity or context. Resolving coreferences is important for extracting correct facts and also understanding and answering user's questions. 

\textbf{3. Entity and relation resolution}: An entity or relational phrase can appear in various surface forms in user's utterance. E.g. entity ``\textit{Obama}" vs "\textit{Barack Obama}" or relation ``\textit{born in}" vs. ``\textit{palace of birth}". Entity and relational phrase resolution and also unseen relation detection are important for optimal knowledge acquisition and inference.

Although (1), (2) and (3) have been studied as independent NLP problems by many existing works, solving them in the conversation modeling context and in lifelong setting and integrating their solutions together to build a holistic knowledge learning system remains a challenging task.

\section{Language Learning in Task-oriented Chatbots}
Task-oriented chatbots (or commonly known as virtual assistants) are Natural Language Interface (NLI) systems that allow users to issue NL commands to the bot  and then the bot interprets the commands and map them into some actions to be executed by the underlying application. Existing methods for building NLIs are of two broad categories. The first category \textbf{(1)} views the process as end-to-end modeling, where a NL command is provided and the system directly outputs the action to be performed. For example, authors of \cite{macmahon2006walk,branavan2009reinforcement,vogel2010learning,misra2017mapping,fried2018speaker,tellex2020robots} have explored deep learning and reinforcement learning to ground NL commands directly into executable actions. The other category \textbf{(2)} focuses on learning a semantic parser to parse the NL command from the user into an intermediate logical form and then, translate the logical form into an executable action in the application \cite{zelle1996learning,artzi2013weakly,andreas2015alignment,zettlemoyer2012learning,DBLP:series/synthesis/2018Li}. In both approaches, the ability to learn previously unknown language expressions and ground them to suitable actions during conversation can greatly improve the performance of NLI systems.

Based on the \textit{various modalities of human-chatbot interactions}, we organize the scope for learning new language expressions in the following two categories:

\begin{itemize}
    \item \textbf{Learning via user demonstrations.} In some cases, NLI systems deployed in practice come with Graphical User Interfaces (GUIs) or remote control facilities to explicitly control  devices apart from controlling them via NL commands. Examples of such systems include task completion robots performing household activities like cleaning robots and personal assistant services integrated with home appliances like Smart TVs, Smart Lights, Smart Speakers, etc. 
    Considering the user has issued an NL command and the bot has failed to execute the intended action, the user may perform the intented action via the GUI or remote control. The bot can record the sequence of executable action(s) performed by the user by accessing the underlying application logs and store the executed APIs as ground truth for the input NL command. The command along with the invoked APIs can serve as labeled examples for learning the command. Related research includes \cite{wang2016learning,wang2017naturalizing}. 
    
   \item \textbf{Learning via multi-turn NL dialogues with the user.} In many cases, demonstration may not be possible either because the user does not know how to do it or no GUI or remote control exists. 
Learning language via multi-turn NL dialogues with the end users may be the only option. For example, the user issued the command ``\textit{turn off the light in the kitchen}" and the bot has failed to execute the intended action. The bot can show/tell a list of \textit{top-k} predicted actions as NL descriptions (as shown below) that can be executed in the current state of the application and asks the user to select the appropriate option from the list. 
    
    \vspace{2mm}
    \begin{quote}
    \small
    \rule{0.9\linewidth}{0.4pt}\\
    \textbf{Bot}: Please choose the correct action option below:\\
    \vspace{-2mm}\\
    \textbf{option-1.}~~ Switch on the light at a given place. \\ 
    \textbf{option-2.}~~ Change the color of light to a given color.\\
    \textbf{option-3.}~~ Switch off the light at a given place. \\
    \rule{0.9\linewidth}{0.4pt}
    \vspace{-1mm}
   \end{quote}
    
    The user can easily select the right option (option-3). The action API [say, \texttt{SwitchOffLight}(\textit{arg}:place)] corresponding to the selected option (here, option-3) can be regarded as the ground truth for the issued NL command, which is to be used as a new example for learning the new language expression. In subsequent turns of the dialogue session, the agent can also ask additional NL questions and show option list to acquire the ground truth values of the arguments of the (ground truth) action API. 

    ~~~ Recently, \citet{mazumder2019building} proposed a method to enable dialogue based language learning in the context of building NLIs. One of the key issues is how to understand paraphrased NL commands from users in order to map a
    user command to a system’s API call. The system aims to learn new paraphrased commands when it has difficulty to understand a user command via an interactive dialogue with the user. In this way, the system becomes more powerful. 
    When the same or a similar NL command is issued by this or another user, the system will have no problem to understand it. In experiments, the authors showed that the command grounding performance improves by 3.6\% when the NLI system is allowed to interactively learn new commands from users compared to that when the language learning is disabled.

\end{itemize}

\section{Learning of Conversation Skills}
A dialogue system can also learn conversation skills to carry out more meaningful and engaging conversations with users. This type of learning is especially important for chit-chat systems so that over time, it can provide more human-like conversation experience to end users. 
Some of the major scopes for learning conversation skills are as follows:

\begin{itemize}
    \item \textbf{Learning user behaviours and preferences}: User's dialogue history is a valuable resource to learn each user's behaviors and preferences in various conversation contexts. Given a conversation context, the chatbot can learn whether a user feels more excited or gets annoyed while conversing on a particular topic, what his/her likes and dislikes are etc. to build the user's behavioral and preference profile. The chatbot can then utilize this user profile knowledge in modeling future conversations to make them more engaging with the user. 
    
    \item \textbf{Learning emotions and sentiments}: Recognizing emotional state \cite{chatterjee2019semeval} and sentiments of the user and leverage it to generate empathetic responses can be useful to building therapeutic chatbots. Some recent works \cite{zhou2018emotional,zhang2017building,sun2018emotional,peng2019topic} have studied the problem of emotional conversation modeling. 
    
    \item \textbf{Modeling situation-aware conversations}: Understanding the situation and spatial-temporal context of a person to decide the conversation strategy is a key characteristic of human conversation process. Continuously learning from the conversation history of the user provides a scope for chatbots to learn user's conversation profile, e.g., what time of a day the user generally likes to talk or remains busy; understanding spatial-temporal context of the user like whether the user is in a meeting or not, etc. can be useful in building situation-aware proactive chatbots and this can improve user's conversation experience.
\end{itemize}

\section{Some Other Challenges}
In this section, we highlight some other challenges, which also present potential research opportunities. One obvious challenge is \textit{few-shot learning} as the ground truth training examples obtained during conversations are scarce. But we will not discuss it here as it is already a well-known problem. Below, we focus on a few other major challenges.  

\vspace{+1.5mm}
\textbf{1. Dealing with Wrong Knowledge from Users.}
As we proposed to learn new knowledge through interactions with the end users, one major challenge is how to deal with the issue of acquiring intentional or unintentional wrong knowledge from end users. For example, while providing demonstration of an action or in a dialogue session with the agent, the user may perform a wrong action for a given input command or provide an incorrect feedback to the agent to erase its old learning. Then the agent may display unintended behaviour, which might even lead to safety issues.

Since chatbots almost always work in a multi-user environment, such issues can be addressed through a \textit{\textbf{cross-verification}} strategy. After acquiring a piece of new knowledge (a new command pattern or action ground truth) in an interaction session, the agent can store these new examples in a unverified knowledge buffer. Next, while interacting with some other users in future sessions to accomplish a related task, it can ask questions to verify the accumulated unverified knowledge. Once a labeled example is verified for $K$ times (from $K$ different random users), the example can be considered as trustworthy and removed from unverified knowledge buffer to be used in learning.

\vspace{+1.5mm}
\textbf{2. Revision of Knowledge.}
Although strategies can be designed to cross-verify any knowledge learned from users, some wrong knowledge will inevitably be learned and stored in the knowledge base. The challenge is how to revise or correct the wrong knowledge once it is detected. This requires a knowledge monitoring system that can detect contradictions in the knowledge base and also knowledge revision method that can revise the wrong knowledge and also all the consequences inferred from it. These are challenging tasks. 

\vspace{+1.5mm}
\textbf{3. Dealing with Safety and Ethical Issues.}
The ability to learn continuously during conversations comes with the problem of abusive language learning from end users. Also, learning user's behavior, situation and emotional profile and using the knowledge in unintended ways can become a risk to user privacy hacking and biased conversation modeling. Thus, \textit{constrained conversational modeling} is needed to prevent unintended sharing and abusive use of user information.

\vspace{+1.5mm}
\textbf{4. Learning New Task Completion Skills from Users.}
Modern task-oriented chatbots are deployed with a finite set of task completion skills which they have been pre-programmed with to perform. Building solutions to enable end users to use natural language dialogues to program their own chatbots and endow them with new skills after deployment will lead to personalization of virtual assistants.  

\section*{Acknowledgments}
This work was supported in part by a research gift from Northrop Grumman and two grants from National Science Foundation: IIS-1910424 and IIS-1838770. We thank former and current students and collaborators, 
Jiahua Chen, Zhiyuan Chen, Sepideh Esmaeilpour, Geli Fei, Wenpeng Hu, Zixuan Ke, Gyuhak Kim, Huayi Li.
Guangyi Lv, Nianzu Ma, Arjun
Mukherjee, Qi Qin, Lei Shu, Hao Wang, Mengyu Wang, Shuai Wang, and Hu Xu, for contributing many ideas on lifelong learning and dialogue systems.

\bibliography{aaai19}
\bibliographystyle{aaai.bst}

\end{document}